\documentclass{article}

    \PassOptionsToPackage{numbers, compress}{natbib}

    \usepackage[final]{neurips_2024_ml4ps}

\usepackage[utf8]{inputenc} 
\usepackage[T1]{fontenc}    
\usepackage{hyperref}       
\usepackage{url}            
\usepackage{booktabs}       
\usepackage{amsfonts}       
\usepackage{nicefrac}       
\usepackage{microtype}      
\usepackage{xcolor}         
\usepackage{forest}
\usepackage{soul}
\DeclareUnicodeCharacter{2212}{-}
\usepackage{multirow}
\usepackage{graphicx}
\usepackage{dcolumn}
\usepackage{bm}
\usepackage{soul}
\usepackage{forest}
\usepackage{booktabs}
\usepackage{amsmath}
\usepackage{subfigure}
\usepackage{tikz}
\tikzset{
    ncbar angle/.initial=90,
    ncbar/.style={
        to path=(\tikztostart)
        -- ($(\tikztostart)!#1!\pgfkeysvalueof{/tikz/ncbar angle}:(\tikztotarget)$)
        -- ($(\tikztotarget)!($(\tikztostart)!#1!\pgfkeysvalueof{/tikz/ncbar angle}:(\tikztotarget)$)!\pgfkeysvalueof{/tikz/ncbar angle}:(\tikztostart)$)
        -- (\tikztotarget)
    },
    ncbar/.default=0.2cm,
}
\tikzset{square left brace/.style={ncbar=0.1cm}}
\tikzset{square right brace/.style={ncbar=-0.1cm}}

\usepackage{enumitem}
\setlist[itemize]{align=parleft,left=0pt..1em}
\title{A Perspective on Symbolic Machine Learning in Physical Sciences}

\author{%
  Nour Makke
  \\
  Qatar Computing Research Institute\\
  Doha, Qatar \\
  \texttt{nmakke@hbku.edu.qa} \\
  \And
  Sanjay Chawla \\
  Qatar Computing Research Institute\\
  Doha, Qatar \\
  \texttt{schawla@hbku.edu.qa} \\
}

\begin{document}

\maketitle

\begin{abstract}
Machine learning is rapidly making its pathway across all of the natural sciences, including physical sciences. The rate at which ML is impacting non-scientific disciplines is incomparable to that in the physical sciences. This is partly due to the uninterpretable nature of deep neural networks. 
Symbolic machine learning stands as an equal and complementary partner to numerical machine learning in speeding up scientific discovery in physics. This perspective discusses the main differences between the ML and scientific approaches. It stresses the need to equally develop and apply symbolic machine learning to physics problems, in parallel to numerical machine learning, because of the dual nature of physics research. 
\end{abstract}

\section{Introduction}

The past decade has witnessed the increasing application of machine learning (ML) methods to natural sciences in general, including physical sciences~\cite{RevModPhys.91.045002}. 
The rise of deep learning (DL)~\cite{leCun} in early 2010 and the remarkable potential of deep neural networks (DNNs) in learning highly predictive models, mainly powered by convolutional~\cite{lecun95} and recurrent~\cite{Sherstinsky_2020} neural networks, emphasized with the ImageNet challenge~\cite{imagenet} and developments in areas such as reinforcement learning~\cite{mnih2013playing}, have boosted the application of artificial intelligence (AI) in nearly all domains and thus reshaped the future of AI. 
The DL revolution was followed by the successful transformer architecture~\cite{DBLP:journals/corr/VaswaniSPUJGKP17}, where the concept of ``attention'' was added into standard NN's architecture to capture long-term correlations between data features. Transformers are the building block of large language models (LLMs), which can learn the context in sequential data without domain-specific knowledge by pretraining on large datasets, thus unlocking another new era of AI. Despite the tremendously evolving AI, yet, most ML-based applications in physical sciences~\cite{RevModPhys.91.045002} focuses on learning non-linear numerical models to accomplish specific tasks (e.g., data analysis, simulations, etc.) toward achieving new findings.
Here arises the question of physicists'  expectations from applying ML and how it could advance physics. Is it simply a set of revolutionary mathematical tools whose performance overcomes classical approaches and thus replaces them (e.g., DL out-performed boosted decision trees for an event-selection task
targeting hypothesized particles from theories beyond the
standard model of particle physics~\cite{Baldi_2014}), or could it be an enabler for data-driven scientific discovery~\cite{Makke2024}? 

It is useful to introduce and define numerical and symbolic ML, and question if only numerical ML is sufficient to make discoveries in physical sciences. 
This paper distinguishes numerical and symbolic ML methods based on their tasks and outputs rather than their inner mechanisms. Numerical ML methods produce numerical outputs, while symbolic ML methods generate symbolic expressions that can directly be interpreted by humans or accomplish pre-defined tasks involving mathematical symbols. Importantly, both approaches rely on numerical optimization of model parameters during training. The distinction between numerical and symbolic ML is independent of NNs' architectures.

The main points of this perspective are (1) that the rate at which ML is applied to physical sciences is relatively slow, this could be mostly due to the uninterpretable (i.e., black-box) nature of DNNs, (2) the application of symbolic ML to physical sciences is at a very early stage compared to numerical ML, and (3) the fundamental connection between experimental and theoretical physics requires to be advancing both pillars in physical sciences at a relatively comparable speed using ML. 

\section{Physical Sciences versus Machine Learning}

The ML methodology focuses on learning (predominantly numerical) models that describe well ``training'' data so that they can produce highly accurate predictions for ``test'' data. There isn't, in principle, any consideration of understanding the models' parameters or weights, leading to a profound difference between the scientific and ML approaches, as summarized in Tab.~\ref{tab:comparison}. 

\begin{table}[]
\centering
\caption{Comparison between Physical Sciences and Artificial Intelligence.}
\begin{tabular}{p{2.5cm}||p{5cm}||p{5cm}}
\toprule
 & \textbf{Physical sciences} & \textbf{Artificial Intelligence} \\
\midrule
Input & Physical Observables and prior knowledge & Data and task description\\
\midrule
Model's expectation & Explain underlying mechanisms in complex systems & Perform the required task with high accuracy \\
\midrule
Ultimate goal & Create, advance and transfer knowledge & Build autonomous systems \\
\midrule
Model's requirement & A physical model is required to (i) Describe observations, (ii) Generalize to other energies, observables, domains, etc., (iii) Connect with existing theories, and (iv) Make predictions.
& An ML model is required to (i) describe well ``training'' data, (ii) perform well on ``test'' data, and (iii) successfully carry on the tasks in an autonomous fashion
\\
\midrule
Success criteria & clear structure, connectivity with previous principles/theories & predictivity 
\\
\midrule
Considerations & parsimony & trend towards large models \\
\midrule
Operation level & the latent structure is the target physical model & the latent structure is conceptual \\
\midrule
Discoveries & Goal & Augmentation (e.g., hypothesis generation)\\
\bottomrule
\end{tabular}
\label{tab:comparison}
\end{table}

The ultimate goal of physical sciences, and generally natural sciences, is to ``understand'' the universe through the naturally occurring mechanisms that come into play. 
Understanding traditionally occurs with progress, which is in turn achieved by (1) discovering new types of phenomena, objects, theories, etc., (2) explaining new discoveries, (3) connecting them with established theories to unify knowledge, and (4) predicting new kinds of phenomena and objects. 
This could be accomplished by learning the latent structure of the world or the physical system we study. 
For example, astrophysicists attempt to understand the internal structure of the Sun or the mechanisms that form planets, and particle physicists investigate the internal structure of the proton, which is by far the most abundant form of matter in our visible Universe, or how elementary particles interact to form the visible matter, or the mechanisms through which elementary particles form composite particles.

Such goals are achieved by learning physical models whose parameters have physical interpretations and are not simply numerical. Learning insights about models' parameters and their dependence upon involved features is a primary task. 
This is related to the dual nature of the physical sciences, where a ``theory'' is a mathematical formulation of a concept that explains ``experimental'' observations.
Here, it is worth noting a few points. First, discoveries in physical sciences could be made either experimentally or theoretically (e.g., a new particle could be observed in an experiment or predicted by a theory). Second, theory and experiment are fundamentally interconnected, so a theory-experiment matching is required for any discovery to be confirmed.
For example, theorists F.~Englert and P.~Higgs predicted the existence of the Higgs boson in 1964~\cite{PhysRevLett.13.321,PhysRevLett.13.508}.  The prediction was only confirmed around $50$ decades later when the new particle's faint signal was observed in 2012 at the large hadron collider  experiments~\cite{higgs,201230} at CERN. 
The Higgs prediction is an ideal example in physical sciences showing (i) how a theory-experiment matching is a key requirement to confirm any prediction, and (ii) how complementary the two pillars are, i.e., developments in experimental particle physics over a few decades were guided by one theoretical prediction, while the development of advanced theories was undergoing.
A few related questions, that would help us shape and demarcate the directions of the research at the intersection between ML and physics, arise:
\begin{itemize}\setlength\itemsep{.1em}
    \item Is ML-based discovery of Higgs from experimental data possible without prior knowledge? 
    \item Is ML capable of discovering Higgs from collected experimental data? 
    \item Is ML capable of making Higgs's theoretical prediction?
\end{itemize}


Deep learning (DL) has proven highly powerful for characterizing numerical data and stands successful in discovering hidden patterns in different types of data in various domains.
The application of DL in physical sciences, however, remains at an early stage compared to others and only pours into numerical operations tasks, where black-box models achieve higher performance than traditional methods at much shorter timescales. 
 In particle physics, there is an increasing amount of research work where ML is deployed in high-energy physics, which is summarized in~\cite{feickert2021livingreviewmachinelearning}. ML algorithms have played a central role in unveiling Higgs's signal, being used in many aspects of data analysis ranging from regular tasks such as classification and tagging to unsupervised anomaly detection. Whereas their application has significantly accelerated the discovery, they were still used as mere tools that outperformed standard approaches to statistical analyses of enormous amounts of data. 
In view of the lack of ML/DL interpretability, which remains a central question in the AI community, the application of black-box models has been slowed down in physical sciences, even being used as tools to accomplish specific tasks. 
A recent position paper~\cite{hogg2024machinelearninggoodbad} elucidates two important and strong statistical biases that are being introduced to the natural sciences by some uses of ML. It discusses the question of whether ML is good or bad for natural sciences.   


A recent trend showed up in the last few years to move from data-centric ML to causal ML by introducing abstract concepts such as graphs, symmetry, symbolic representations, etc. These concepts are more connected to fundamental physics and extend physicists’ expectations to a new horizon. Numerical ML is reaching a highly advanced status, and it is expected to be more impactful in experimental physics. If theoretical developments do not occur at a similar timescale as experimental ones in the artificial intelligence age, discoveries in experimental physics will remain disputable for a long time. Here emerges symbolic ML as an equal partner to numerical ML. 
The application of symbolic ML in physics remains at an early stage. A search in the inspire-hep database shows that research works on symbolic ML in physics were initiated a few years ago (i.e., 2015-2024), compared to numerical ML, which started back a few decades ago (i.e., 1972-2024).



\section{Symbolic Machine Learning}

\begin{figure}[t]
\centering
\includegraphics[width=.55\linewidth]{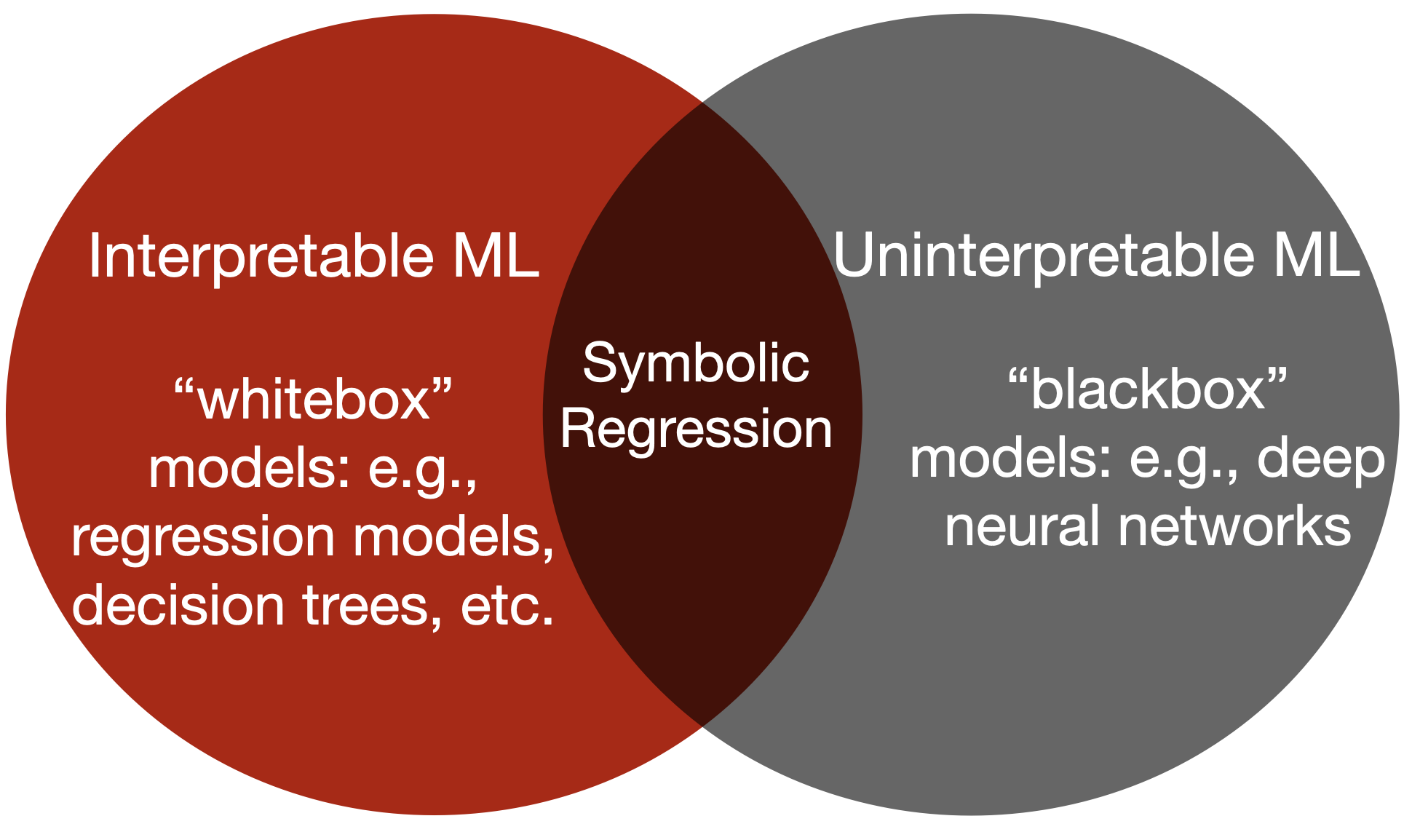}
\caption{Symbolic Regression~\cite{Makke2024}: The art of learning white-box models using black-box models.}
\label{fig:2}
\end{figure}

Two exploratory paths can be undertaken in symbolic ML, either for use in symbolic manipulation tasks in mathematical physics or for learning analytical models from numerical data points. One particularly important application of symbolic ML is symbolic regression~\cite{Makke2024}, aiming to learn mathematical equations from numerical data points. Symbolic regression (SR) can be regarded as the intersection between interpretable and uninterpretable ML (See Fig.~\ref{fig:2}), i.e., SR can be used to learn by design interpretable models using black-box models. SR is growing in popularity because it promotes interpretability and transparency. However, it remains at a very early stage of development compared to numerical ML.
Recent applications of symbolic ML fold in symbolic manipulation tasks in the study of the scattering of elementary particles in high-energy physics. First is the computation of the squared scattering amplitude~\cite{symba}, a key element of the cross-section calculation, and the second~\cite{polylog} is the calculation of polylogarithmic functions that arise when computing the cross-section at high orders of the perturbation theory (so-called Feynman integrals). 

\subsection{Symbolic Regression}

Symbolic regression (SR) is an area of AI for inferring symbolic expressions from numerical data. 
Unlike ``black-box'' models, SR provides crucial insights for advancing scientific understanding by developing ``white-box'' models. 
The concept of SR is as old as Kepler's discovery~\cite{kepler} of elliptic orbits in the 16$^{\text{th}}$ century. Modern SR emerged back in 1970 and has been explored for decades in the context of scientific discovery first by developing BACON~\cite{bacon1, 10.5555/29379} followed by COPER~\cite{coper}, FAHRENHEIT~\cite{fahrenheit1,fahrenheit2} and LAGRANGE~\cite{lagrange}. SR has since been developed by the genetic and evolutionary computation community. Although SR has been mainly developed and promoted as an AI tool for automated scientific discovery, it has not been adopted in the physical sciences. This could be explained by the limitations encountered in both areas. For example, the accuracy reached in current experimental measurement is in no way comparable to the accuracy that was reachable a few decades ago, and SR methods were incapable of processing high-dimensional datasets in addition to limited computation resources in light of the exponentially growing size of the search space in SR.

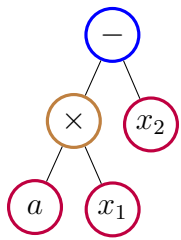
\begin{figure}[t]%
\centering
\begin{forest}
for tree = {circle, draw,
        minimum size=2.em,
        inner sep=0.5pt,
        font=\large,
        l sep=4mm,s sep=3mm
        },
        rbcirc 2/.style={%
            circle,
            fill=white,
            draw=blue,
            very thick,
        },
        rbcirc 3/.style={%
            circle,
            fill=white,
            draw=brown,
            very thick,
        },
        rbcirc 4/.style={%
            circle,
            fill=white,
            draw=purple,
            very thick,
        },   
[$-$,rbcirc 2[$\times$,rbcirc 3[$a$, rbcirc 4][$x_1$, rbcirc 4]][$x_2$, rbcirc 4]]
\end{forest}%
\caption{Expression-tree structure of the mathematical equation $f(x_1,x_2)=ax_1-x_2$. Root (blue), internal (brown), and terminal (purple) nodes are differently colored.}
\label{fig:treestructure}
\end{figure}

Following the DL revolution, SR is re-emerging as a potential candidate to overcome the interpretability issue in AI models and automate discovery in sciences. SR methods have evolved significantly from traditional search-based approaches (e.g., heuristic search and evolutionary algorithms) to modern learning-based (e.g., transformer-based models) and hybrid techniques. A full review on SR can be found in~\cite{Makke2024,makkekdd,livingreview}. 
A powerful representation of mathematical equations is the unary-binary tree, where operators appear as internal nodes and operands as leaves (terminal nodes), as illustrated in the example of Fig.~\ref{fig:treestructure}. This tree structure enables the conversion of any equation into a sequence of mathematical symbols (using either prefix or postfix notation), making it possible to leverage sequence-to-sequence models for learning symbolic expressions similar to learning linguistic expressions, in light of the great power of transformer NNs. For example, the function $f(x_1,x_2)=ax_1-x_2$ is equivalent to $\{-,\times,a,x_1.x_2\}$ in the prefix notation, commonly referred to as the Polish notation~\cite{polishnotation}.

SR is particularly suited for application in physical sciences by 
generating parsimonious models that balance accuracy and simplicity in a human-understandable form, thus allowing physicists to understand the mathematical structure behind the studied phenomena. On the other way, the physical sciences represent a rich source of data, either synthetic or experimental, that can be used to train ML-based SR methods. Physics data is a valuable input to test the credibility of SR methods, given that physical equations are causal relations and physical variables have values in specific numerical ranges and cannot take any arbitrary values.

The application of SR in physical sciences remains limited despite the significant advances in developing DNN-based SR methods. Moreover, most SR applications use simulated data, e.g.~\cite{Ashhab2024}, and very few applications use experimental (noisy) data, which are particularly important to test the credibility of SR to automate discovery and to potentially understand the current limitations of SR methodologies. A recent SR application~\cite{cranmersolar} uses astronomical data (i.e., positions and velocities of planets in the solar system) collected by NASA over thirty years and successfully learns Newton's law of gravity and the planets' masses. 
The key success of this work lies in training a graph neural network (GNN) on observed data, where nodes represent planet characteristics and edges capture their interactions, and then applying SR to the GNN's edges to derive the equation governing planetary interactions. Here, a ``black-box'' model is used to learn the pattern in real data, which is subsequently transformed into a human-interpretable expression through SR. 
This two-step approach proves particularly valuable when SR struggles to infer meaningful models from real (by default noisy) data whose dynamic range spans several orders of magnitude. In such cases, DNNs are leveraged to learn accurate but uninterpretable models. This study presents an approach to follow for challenging data, but also critically highlights significant limitations in current SR methods that demand future research and refinement.

The study in~\cite{10.1093/pnasnexus/pgae467} presents another successful application of SR to differential hadron distributions measured by a high-energy physics experiment at CERN. This study is the first to use such a high-statistics measurement of differential hadron distributions, i.e., $\mathcal{M}(\mathrm{x})$ with $x\in\mathbb{R}^4$, and directly applies SR on real data, in contrast to the previous study~\cite{cranmersolar}. The learned function closely resembles the well-known Tsallis function, which is traditionally used to fit experimental hadron distributions measured across various high-energy physics experiments and different center-of-mass energies. 
Another successful SR application is presented in~\cite{makke_ml4ps_article}, where SR is used to learn a functional form of the universal fragmentation functions (FFs) from experimental hadron multiplicities. FFs represent a key ingredient to calculate the cross-section of hadron production in all high-energy physics experiments producing final-state hadrons. Traditionally, FFs are determined by fitting the parameters of a pre-assumed functional form to experimental data. This study is the first to infer a functional form of FFs, without any constraints, which is found to closely resemble the Lund string model of the hadronization mechanism. This finding represents the first experimental confirmation of the Lund model, which is widely used in high-energy physics simulations. 

These recent applications of SR to experimental data serve multiple purposes: first, they systematically identify the limitations of current SR methods, providing a roadmap for future methodological improvements; second, they evaluate the most effective SR approaches for physics applications; and ultimately, they demonstrate SR's credibility to learn accurate and concise analytical models from experimental data, thus positioning SR as a potential tool automated scientific discovery. In addition, the effective deployment of SR demonstrates that human intervention and prior knowledge of data are essential for producing scientifically valid and robust results

\section{Conclusions}

Bridging fundamental physics and machine learning implies that both theoretical and experimental approaches are equally involved. If numerical machine learning is a key player in the future of experimental physics, symbolic machine learning will certainly be an equal partner in theoretical and phenomenological physics.  

Symbolic regression, among other ML techniques, emerges as a potential candidate for inferring analytical models from numerical data. It is particularly suited for physical sciences since physics is expressed by symbolic representations. 
Significant effort is needed from physicists to develop and deploy symbolic ML and SR in physics problems. The first applications of SR applications on experimental data in physical sciences are especially promising. However, they also point to many limitations that need to be addressed for fruitful applications of SR toward new physics discoveries. 

\section{Acknowledgments}

We thank the referees at the Machine Learning and the Physical Sciences (ML4PS) Workshop at the Thirty-eight Conference on Neural Information Processing Systems (NeurIPS 2024) for comments and feedback on this work.




\bibliography{apssamp.bib}

\end{document}